\RequirePackage{fix-cm}
\documentclass[runningheads]{llncs}
\usepackage{graphicx}
\usepackage{lineno,hyperref}
\modulolinenumbers[5]

\usepackage{graphicx}
\usepackage{amssymb}
\usepackage{amsmath}
\usepackage{adjustbox}
\usepackage{makeidx}  % allows for indexgeneration
\usepackage{array}
\usepackage[caption=false]{subfig}
\usepackage{booktabs}
\usepackage{textcomp}
\usepackage{todonotes}
\usepackage{wrapfig}
\usepackage[shortcuts,acronym]{glossaries}

\newacronym{RCNN}{RCNN}{Mask-R Convolutional Neural Network}
\newacronym{CNN}{CNN}{Convolutional Neural Network}
\newacronym{BCL}{BCL}{Bilateral Convolution Layer}
\newacronym{USM}{USM}{Unsharp Masking}
\newacronym{PC}{PC}{PlantCLEF}
\newacronym{NLM}{NLM}{Non Local Means}
\newacronym{DNLM}{DNLM}{Deceived Non Local Means}
\newacronym{SVM}{SVM}{Support Vector Machine}
\newacronym{LoG}{LoG}{Laplacian of Gaussian}
\newacronym{DoG}{DoG}{Difference of Gaussians}
\newacronym{SP}{SP}{Salt and Pepper}
\newacronym{RPN}{RPN}{Region Proposal Network}
\newacronym{ReLU}{ReLU}{Rectified Linear Unit}
\newacronym{FCN}{FCN}{Fully Convolutional Network}
\newacronym{ROI}{ROI}{Region of Interest}
\newacronym{DDT}{DDT}{Deep Distance Transformer}
\newacronym{Top-WS}{Top-WS}{Watersheds instance segmentation as top-model}
\newacronym{Top-RPN-WS}{Top-RPN-WS}{Region proposal network and watersheds top-model}
\newacronym{Top-Unet3}{Top-Unet3}{U-net top-model}
\newacronym{Top-Unet-ML}{Top-Unet-ML}{U-net with modified loss function top-model}
\newacronym{U-net3}{U-net3}{U-net for three classes}
\newacronym{Unet-ML}{Unet-ML}{U-net with the proposed modified loss function}
\newacronym{DTGT}{DTGT}{Distance Transform Ground-truth}
\newacronym{BTGT}{BTGT}{Border Transform Ground-truth}
\newacronym{WDMC}{WDMC}{Weighted Dice Multiclass Coefficient}
\newacronym{BDE}{BDE}{Boundary Displacement Error}
\newacronym{MAE}{MAE}{Mean Absolute Error}
\newacronym{MSE}{MSE}{Mean Squared Error}

\begin{document}

\title{Enforcing Morphological Information in Fully Convolutional Networks to Improve Cell Instance Segmentation in Fluorescence Microscopy Images\thanks{This work is partially supported by the following Spanish grants: TIN2016-75097-P, RTI2018-094645-B-I00 and UMA18-FEDERJA-084. All of them include funds from the European Regional Development Fund (ERDF). The authors acknowledge the funding from the Universidad de M\'alaga.}}

\titlerunning{Image analysis by ensembles of convolutional neural networks}        

\author{Willard Zamora-C\'ardenas$^{1}$, Mauro Mendez$^{1}$, Saul Calderon-Ramirez$^{1,2}$, Martin Vargas$^{1}$, Gerardo Monge$^{1}$, Steve Quiros$^{3}$, David Elizondo$^{2}$, Jordina Torrents-Barrena$^{4}$ and Miguel A. Molina-Cabello$^{5,6}$}

%\author{author1 \and author2 \and ... }

\authorrunning{Zamora-C\'ardenas et al.} % if too long for running head

% \institute{Miguel A. Molina-Cabello \at Department of Computer Languages and Computer Science\\
% University of Malaga\\
% Bulevar Louis Pasteur, 35.\\
% 29071 M\'alaga. Spain.\\
% \email{miguelangel@lcc.uma.es}\\
% }

\institute{$^{1}$Computing School, Costa Rica Institute of Technology, Costa Rica\\
$^{2}$Department of Computer Technology, De Montfort University, United Kingdom\\
$^{3}$Tropical Diseases Research Center, Microbiology Faculty. University of Costa Rica, Costa Rica\\
$^{4}$Department of Computer Engineering and Mathematics, Rovira i Virgili University, Spain\\
$^{5}$Department of Computer Languages and Computer Science, University of Malaga, Spain\\ 
$^{6}$Instituto de Investigación Biomédica de Málaga – IBIMA, Spain\\
E-mail: sacalderon@itcr.ac.cr}

\date{Received: date / Accepted: date}

\maketitle

\begin{abstract}

Cell instance segmentation in fluorescence microscopy images is becoming essential for cancer dynamics and prognosis. Data extracted from cancer dynamics allows to understand and accurately model different metabolic processes such as proliferation. This enables customized and more precise cancer treatments. However, accurate cell instance segmentation, necessary for further cell tracking and behavior analysis, is still challenging in scenarios with high cell concentration and overlapping edges. Within this framework, we propose a novel cell instance segmentation approach based on the well-known U-Net architecture. To enforce the learning of morphological information per pixel, a deep distance transformer (DDT) acts as a back-bone model. The DDT output is subsequently used to train a top-model. The following top-models are considered: a three-class (\emph{e.g.,} foreground, background and cell border) U-net, and a watershed transform. The obtained results suggest a performance boost over traditional U-Net architectures. This opens an interesting research line around the idea of injecting morphological information into a fully convolutional model.

\end{abstract}

\section{Introduction}
\label{sec:introduction}
The application of new image processing techniques is a burgeoning trend in life sciences such as biology, chemistry, medicine, among others. Their implementation includes object measurement, 3D space exploration (\emph{e.g.,} magnetic resonance / positron emission tomography) for surgical planning, dynamic process analysis for time-lapse imaging in cell growth, movement and proliferation \cite{mahesh2011fundamentals,calderon2018automatic,calderon2016first,calderon2015dewaff,saenz2015deceived}, detection and classification of blood cells  \cite{molinacabello2018blood}, cancer diagnosis, histopathology and detection of multiple diseases \cite{morgan1989flow, puig2020assessing,alfaro2019brief,calderon2021improving,calderon2021improvingIEEEAccess,bermudez2020first,oala2020ml4h}. 

%cite{morgan1989flow, oates2009hoechst, qi2012robust, molinacabello2019optimization, smith1988validation}. 

Accurate cell segmentation is crucial for robust heterogeneous cell dynamics quantification (\emph{e.g.,} mitotic activity detection), tracking and classification, which are often implemented as subsequent higher-level stages. For instance, intra-tumoral heterogeneity contributes to drug resistance and cancer lethality \cite{mcgranahan2017clonal}. In cancer research, cell biologists aim to monitor single-cell changes in response to chemotherapies. Indeed, given the relevance of malignant cell proliferation, the aforementioned changes need to be rigorously tracked along the progeny of cancer cells through time-lapse microscopy. Manual cell segmentation is time-consuming, prone to human subjective variation and biased by medical devices, making (semi-)automatic cell segmentation approaches appealing.

Fully automatic segmentation of cell instances is widely tackled in the literature (see Section \ref{Previous_work}). Image segmentation is defined as the assignment of a class or label to a pixel (\emph{i.e.,} pixel-wise classification). More specifically, the problem of assigning one out of multiple classes to a pixel is known as semantic segmentation. As for the problem of assigning a different label to a pixel for different instances of a semantic class, it is known as instance segmentation \cite{long2015fully}. The latter is more challenging and provides useful information in several application domains \cite{bai2017deep}.
%\cite{bai2017deep, gupta2014learning, ren2017end, romera2016recurrent}.

Cell counting solutions, a common application for automatic image analysis, often avoids instance segmentation. For example, Weidi Xie \emph{et al.} \cite{xie2018microscopy} used counting estimation in cell clusters and a \ac{CNN} for cell density estimation. However, precise instance detection is essential in cell tracking and individual cell behavior analysis \cite{calderon2015dewaff}. 

Cell instance segmentation different challenges such as intensity saturation and overlapping edges \cite{bai2017deep} that can hinder segmentation accuracy. Noise and poor contrast caused by variable molecule staining concentration are common drawbacks in cell imaging. Furthermore, the low number of manually annotated samples is a frequent limitation, specially for \ac{CNN}-based approaches, as expert\'s knowledge is crucial to generate ground-truth data. Dealing with unbalanced datasets is another recurrent issue, since foreground and overlapping cell pixels are much lower than background pixels \cite{guerrero2018multiclass}. 

In this paper, we devise a novel \ac{CNN} architecture based on the popular U-Net \cite{ronneberger2015u} and the distance transform morphological operator \cite{grevera2007distance} to jointly improve instance segmentation accuracy. Section \ref{Previous_work} presents related work in the field of cell instance segmentation. Later, the proposed method is detailed in Section \ref{secProposedmethod}. Section \ref{Materials} briefly describes the methods and datasets used to subsequently define the experiments. Results are discussed in Section \ref{secExperiments}. Finally, conclusions and future lines of research are described in Section \ref{secConclusions}.

%\input{related}
% Related Work
\section{Related Work}
\label{Previous_work}
Image segmentation is a well-studied and documented problem among the image processing, pattern recognition and machine learning communities. Several methods have been proposed for cell segmentation \cite{meijering2012cell}, which rely on thresholding \cite{phansalkar2011adaptive} or active contours \cite{zamani2006unsupervised} approaches, among others. 

Additionally, the \ac{FCN} \cite{long2015fully} inspired U-net, which  feeds up-sampling layers with the output of different down-sampling layers, enforcing global information at fixed scales \cite{ronneberger2015u}.

According to different approaches to mixing local and global information, in this work, we combine the feed subsequent layers with data from previous convolutions at different scales and pre-process morphological operators directly in the model input. This way, our proposal enforces both local and global information by learning morphological transformations from the data.

% Proposed method
\section{Proposed method} 
\label{secProposedmethod}
\begin{figure}[t]
\centering
\includegraphics[width=0.60\textwidth]{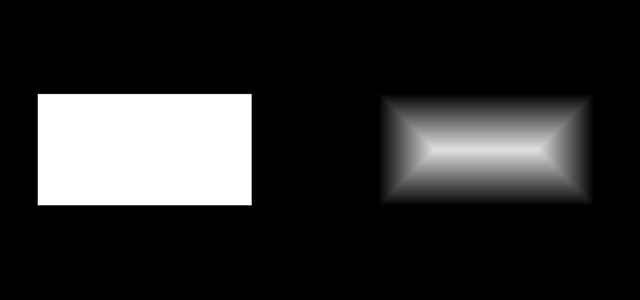}
\caption{Rectangular mask (\emph{left}) and its original distance transform (\emph{right}). The inverse of the distance transform is subsequently computed to provide high intensities to border pixels.}
\label{fig:DT}
\end{figure}

Our segmentation method is based on the well-known U-net architecture \cite{ronneberger2015u}. We propose to enforce cell morphological information in a \ac{CNN} by estimating the inverse of the distance transform. Note that the original distance transform calculates the closest Euclidean distance to a background pixel for each foreground pixel. Figure \ref{fig:DT} shows this transformation applied to a rectangular mask, where border pixels have lower intensities. Differently, the inverse of the distance transform is thus computed to provide high intensities to border pixels.
%\cite{fabbri20082d}

% Figure 1

Specifically, an encoder-decoder classifier with a \ac{ReLU} activation is employed. This classifier is fed with fluorescence microscopy images of cells. The proposed Euclidean distance estimator learns the inverse distance transform (\ac{DDT} or backbone-model) of the instance as a binary mask of the input image. The \ac{DDT} encourages the U-net to learn morphological information from the image.  

The following top-models are integrated in the \ac{DDT} back-model:

\begin{enumerate}
 %    \item \ac{DDT} + Watershed: The \ac{DDT} output is the input of the watershed instance segmentation algorithm. All seeds are initialized by performing morphological dilation and erosion to the \ac{DDT} output. The distance transform is commonly used to create the seeds for the watershed algorithm \cite{hamarneh2009watershed}. This top-model avoids an expensive training.
     
    \item DDT + UNet$_{1}$: The \ac{DDT} output is fed to a three-class (\emph{i.e.}, foreground, background or border pixels) predictor using the \ac{BTGT}.
% (see Section \ref{Materials} and Figure \ref{fig:unet_no_dtt})
    
    \item DDT + UNet$_{2}$: Considering that DDT + UNet$_{1}$ could ignore relevant texture information, an extra pipeline with two information channels is included. The first and second channels are the \ac{DDT} output and the original image, respectively.
\end{enumerate}

Table \ref{tab:params} shows the parameters used to train our proposed U-Net based pipelines with an average log of the training and testing times in minutes per epoch.

\begin{center}
\begin{table}[t]
\centering
\caption{Fine-tuned parameters for all models.\\}
%\resizebox{1\textwidth}{!}{
\begin{tabular}{c|c|c|c|c}
\textbf{}                                                                  & \textbf{UNet$_{1}$} & \textbf{DDT} & \textbf{DDT + UNet$_{1}$} & \textbf{DDT + UNet$_{2}$} \\ \hline
\textit{\begin{tabular}[c]{@{}c@{}}Training Time\\ (m/epoch)\end{tabular}}   & $\sim$3.64        & $\sim$3.61   & $\sim$10.50        & $\sim$7.53                   \\
\textit{\begin{tabular}[c]{@{}c@{}}Prediction Time\\ (m/epoch)\end{tabular}} & $\sim$0.35        & $\sim$0.34   & $\sim$1.42         & $\sim$0.66                   \\
\textit{Batch Size}                                                        & 10                & 10           & 10                 & 10                           \\
\textit{Training Size}                                                     & 3690              & 3690         & 3690               & 3690                         \\
\textit{Validation Size}                                                   & 930               & 930          & 930                & 930                          \\
\textit{Learning Rate}                                                     & 0.001             & 0.001        & 0.001              & 0.001                        \\
\textit{Loss Function}                                                     & CE                & MAE          & CE                 & CE                           \\
\textit{Optimizer}                                                         & Adam              & Adam         & Adam               & Adam                        
\end{tabular}
%}
\label{tab:params}
\end{table}
\end{center}

An advantage of our approach over \cite{guerrero2018multiclass} is the stability and speed provided during the training stage. In our experiments, the loss functions proposed by \cite{guerrero2018multiclass} hinders the model convergence. Our methodology is similar to \cite{decenciere2018dealing} as morphological pre-processing is also incorporated. Differently, the presented model learns enriched image textures using pixel-wise labeled data. Instead of performing morphological operations directly in the input data \cite{decenciere2018dealing}, we train an U-net architecture with the distance transform of the ground-truth, thus allowing the model to estimate the same transform for all unlabeled samples.

% Materials
\section{Datasets}
\label{Materials}
The \emph{Broad Bioimage Benchmark Collection BBBC006v1 dataset} \cite{ljosa2012annotated} is used in this study. It consists of Human U2OS cells marked with Hoechst 33342. The Hoechst 33342 is a cellular marker widely employed to stain genomic DNA in fluorescence microscopy images. Due to its staining prowess, the Hoechst 33342 is commonly used to visualize nuclei and mitochondria. It is also widely used in diagnostics, apoptosis,  single nucleotide polymorphism, nuclei acid quantification, epilepsy and cancer behavior analysis \cite{sabnis2010handbook}.

\begin{figure}[t]
\centering
\includegraphics[width=0.50\textwidth]{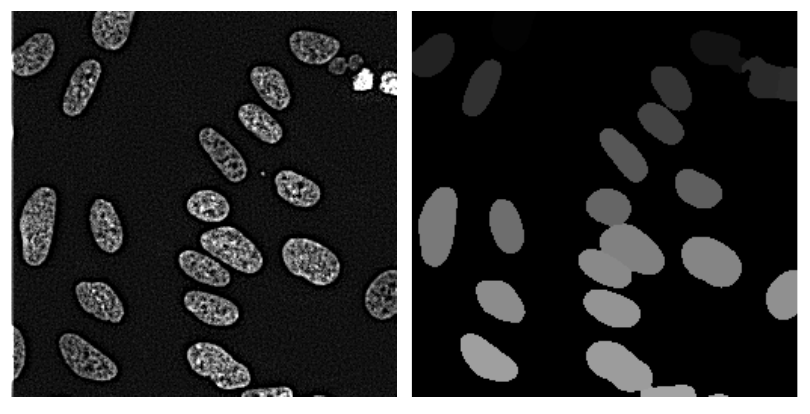}
\caption{\emph{BBBC006} original image with enhanced contrast (\emph{left}) and its corresponding ground-truth (\emph{right}).}
\label{fig:original_and_gt}
\end{figure}

This dataset is composed by 768 images with a resolution of 696$\times$520 pixels in 16-bit TIF format. Each sample includes the corresponding ground-truth image, encoding background pixels with 0. Foreground pixels use different values for labeling each cell instance. To train our models, all samples were tiled to 256$\times$256 pixels (6 tiles per sample = 4608 images). Figure \ref{fig:original_and_gt} shows an original image of the dataset, with its corresponding ground-truth and a copy of it with enhanced contrast to visualize the image content.

% Figure 3

\begin{figure}[t]
\centering
\includegraphics[width=0.75\textwidth]{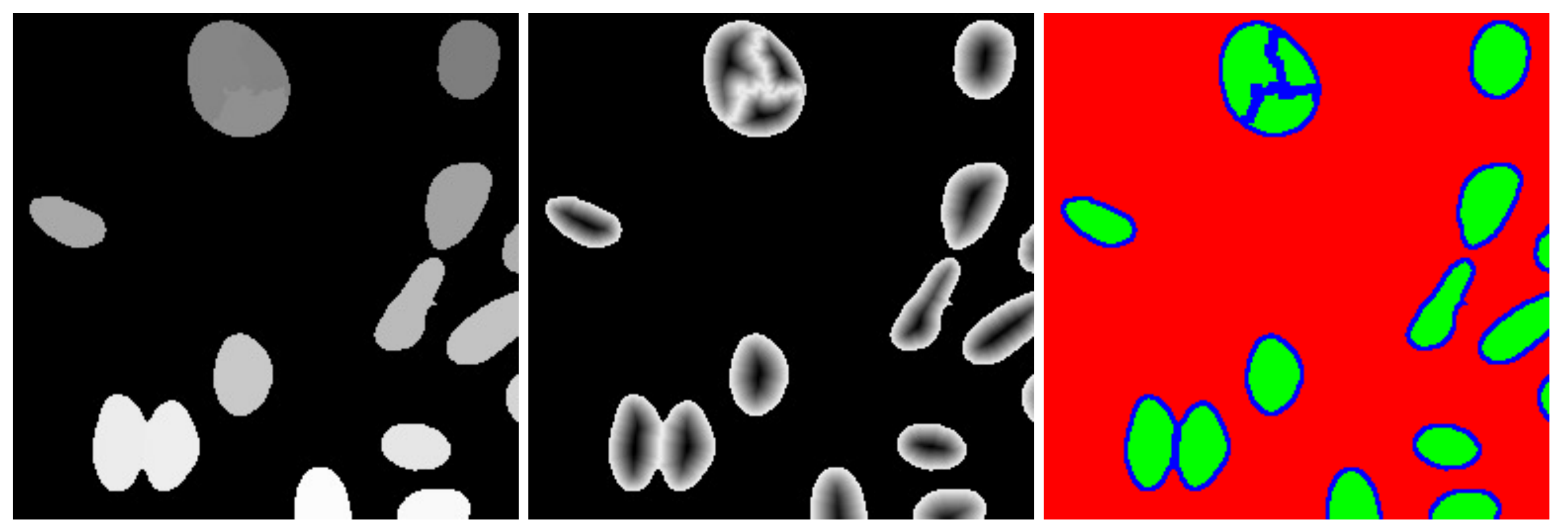}
\caption{Original ground-truth image (\emph{left}), \ac{DTGT} image (\emph{middle}) and \ac{BTGT} image (\emph{right}).}
\label{fig:original_dtgt_btgt}
\end{figure}

Two transformations were also applied to the ground-truth. The first transformation referred to the distance transform (\ac{DTGT}), while the second generated information corresponding to the border between instances (\ac{BTGT}). Specifically, \ac{BTGT} marked pixels as borders if their 3$\times$3 neighborhood contained elements from the outline of another instance. Figure \ref{fig:original_dtgt_btgt} displays the original ground-truth, \ac{DTGT} and \ac{BTGT} images.

% Figure 5

% JORDINA: https://arxiv.org/ftp/arxiv/papers/1803/1803.10829.pdf

% Experiments and Results 
\section{Experiments and Results}
\label{secExperiments}
We used a vanilla three-class U-net architecture to assess the \ac{DDT} improvement over the baseline classification. This model was trained with both the original dataset and the \ac{BTGT}.
%(see Figure \ref{fig:unet_no_dtt}).

% % Figure 2
% \begin{figure*}[t]
% \centering
% \includegraphics[width=0.9\textwidth]{Figures/Figures2-3-6.pdf}
% \caption{DDT back-bone model (\emph{top}), DDT + U-Net$_{1}$ model (\emph{middle}), and U-Net$_{1}$ baseline (\emph{bottom}).}
% \label{fig:unet_no_dtt}
% \end{figure*}

The \ac{DDT} was trained for 50 epochs through the \ac{MAE} loss function. Empirical tests demonstrated that the \ac{MAE} provides higher accuracy than the \ac{MSE}. The U-net$_{1}$ (baseline), DDT + UNet$_{1}$ and DDT + UNet$_{2}$ models were trained end-to-end using the same training observations, and the cross-entropy loss. As aforementioned, the BBBC006v1 dataset was utilized to train both U-net$_{1}$ (baseline) and \ac{DDT} models. The \ac{DDT} output was automatically fed into the DDT + UNet$_{1}$ / UNet$_{2}$ top-models. The DDT + UNet$_{2}$ approach used both feature extractors to enforce texture and morphological information. A repository with the code can be found in its website\footnote{\url{https://github.com/wizaca23/BBBC006-Instance-Segmentation}}.

% The \ac{DDT} output was automatically fed into the DDT + Watershed and DDT + UNet$_{1}$ / UNet$_{2}$ top-models

Inverse distance transformed images were normalized within the range \{0, 1\}. All models were trained using 5-fold cross validation. The number of training and testing images for each fold was 3687 and 921, respectively. %The DDT + Watershed model also took advantage of the \ac{DDT} and was executed once due to the deterministic definition of the watershed seeds.

Edge estimation is crucial for cell instance segmentation, thus specific border accuracy estimation metrics were selected. The \ac{BDE} \cite{unnikrishnan2007toward} averages the displacement error of boundary pixels between two images, in which lower values indicate better matches. The \ac{WDMC} relies on the Sørensen–Dice similitude index \cite{bernard2009variational} (overlapping), so values closer to 1 denote greater similitude. The \ac{WDMC} fine-tuning was conducted by calculating each class separately and performing a weighted sum. In this work, more relevance was given to the correct prediction of border pixels using the following weights found experimentally: $Foreground:0.3$, $Background:0.3$ and $Border:0.4$.

Table \ref{tab:res} shows a comparison of the tested pipelines. The average and standard deviation are the best values for each metric. Results show that the DDT + UNet$_{2}$ model yields a slightly higher accuracy when compared to the UNet$_{1}$ (baseline) model, for both \ac{BDE} and \ac{WDMC} measures. These metrics demonstrated to be more sensitive to edge estimation performance, in spite to more common metrics as the F1-score, and intersection over the union in our empirical tests. Table \ref{tab:scores} shows the average precision, recall and F1 score of each of the proposed pipelines.

The DDT + UNet$_{1}$ model performs slightly better than the UNet$_{1}$ (baseline) according to the \ac{WDMC}, but it has a lower \ac{BDE} accuracy. The DDT + UNet$_{2}$ model outperforms the others, which suggests that the combination of texture information and inverse distance transform improves the overall model performance.

% Table 1
\begin{table}[t]
\centering
\caption{Segmentation accuracy using border displacement error.\\}
\begin{tabular}{c|c|c|c}
\textbf{Pipeline}   & \textbf{BDE}                                                   & \textbf{WDMC}                                                  & \textbf{F1}                                                    \\ \hline
UNet$_{1}$ (baseline) & \begin{tabular}[c]{@{}c@{}}$0.628\pm0.199$\end{tabular}  & \begin{tabular}[c]{@{}c@{}}$0.930\pm0.015$\end{tabular}  & \begin{tabular}[c]{@{}c@{}}$0.946\pm0.006$\end{tabular} \\
%DDT + Watershed & $4.9718$                                                       & 0.8070                                                         & -                                                              \\
DDT + UNet$_{1}$ & \begin{tabular}[c]{@{}c@{}}$0.821\pm0.149$\end{tabular}  & \begin{tabular}[c]{@{}c@{}}$0.924\pm0.006$\end{tabular} & \begin{tabular}[c]{@{}c@{}}$0.922\pm0.010$\end{tabular} \\
DDT + UNet$_{2}$ & \begin{tabular}[c]{@{}c@{}}$0.614\pm0.215$\end{tabular} & \begin{tabular}[c]{@{}c@{}}$0.938\pm0.004$\end{tabular}   & \begin{tabular}[c]{@{}c@{}}$0.938\pm0.004$\end{tabular}
\end{tabular}
\label{tab:res}
\end{table}

\begin{table}[t]
\centering
\caption{Precision, recall and F1 scores for each U-Net pipeline.\\}
\begin{tabular}{c|c|c|c}
\multicolumn{1}{l|}{\textbf{Pipeline}} & \multicolumn{1}{c|}{\textbf{UNet$_{1}$}}                           & \multicolumn{1}{c|}{\textbf{DDT + UNet$_{1}$}}                          & \multicolumn{1}{c}{\textbf{DDT + UNet$_{2}$}}                \\ \hline
Precision                              & \begin{tabular}[c]{@{}c@{}}0,944$\pm $0,008\end{tabular} & \begin{tabular}[c]{@{}c@{}}0,923$\pm $0,013\end{tabular} & \begin{tabular}[c]{@{}c@{}}0,931$\pm $0,006\end{tabular} \\
Recall                                 & \begin{tabular}[c]{@{}c@{}}0,950$\pm $0,007\end{tabular} & \begin{tabular}[c]{@{}c@{}}0,928$\pm $0,013\end{tabular} & \begin{tabular}[c]{@{}c@{}}0,948$\pm $0,002\end{tabular} \\
F1                                     & \begin{tabular}[c]{@{}c@{}}0,946$\pm $0,006\end{tabular} & \begin{tabular}[c]{@{}c@{}}0,922$\pm $0,010\end{tabular} & \begin{tabular}[c]{@{}c@{}}0,938$\pm $0,004\end{tabular}
\end{tabular}
\label{tab:scores}
\end{table}

To assess the statistical significance of the results, we performed a Wilcoxon matched-pairs test. It helps to corroborate the similarity or difference between the UNet$_{1}$ (baseline) model and DDT + UNet$_{1}$ / UNet$_{2}$ model performance. Specifically, we generated 26 random samples from a set of 921 testing images (each with 300 images). Testing images were extracted from the same validation fold, so they were not used for training. Note that we selected the fold where both models performed better. The Wilcoxon test was performed for the \ac{BDE}, \ac{WDMC}. The null hypothesis was formulated as follows: one of the models performs better than the other. Table \ref{tab:wilcoxon} shows the p-values computed for each metric with $0.05$ of significance. They indicate that there is a significant difference between both models with $p<0.05$. The \ac{BDE} is close to be not statistically conclusive, however the \ac{WDMC} presents a more overwhelming statistical difference.  

% Table 2
\begin{table}[t]
\centering
\caption{p-values of the Wilcoxon test.\\}
\begin{tabular}{l|l|l|l}
    & \textbf{BDE} & \textbf{WDMC} & \textbf{F1} \\ \hline
        \textbf{P-value} & 0.02621 & $1.3261 \times 10^{-05}$ & $2.3498 \times 10^{-05}$
%    \textbf{P-value} & 0.02621 & 0.000013261 & 0.000023498
\end{tabular}
\label{tab:wilcoxon}
\end{table}

% Discussion and Conclusions
\section{Discussion and Conclusions}
\label{secConclusions}
\begin{figure}[t]
\centering
\includegraphics[width=0.25\textwidth]{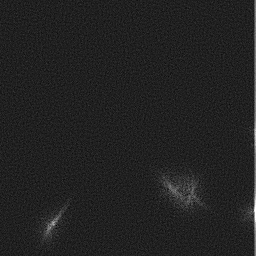}
\includegraphics[width=0.25\textwidth]{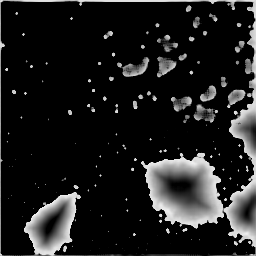}
\includegraphics[width=0.25\textwidth]{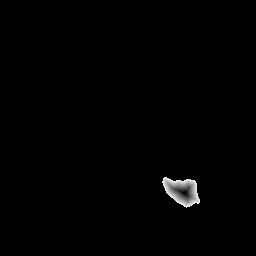}
\caption{Original image with enhanced contrast (\emph{left}), prediction output (\emph{middle}), and current \ac{DTGT} (\emph{right}).}
\label{fig:Back-Model-Noise}
\end{figure}

\begin{figure}[t]
\centering
\includegraphics[width=0.24\textwidth]{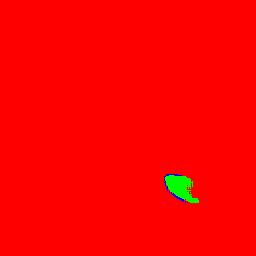}
\includegraphics[width=0.24\textwidth]{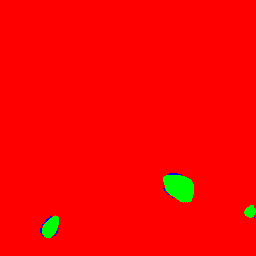}
\includegraphics[width=0.24\textwidth]{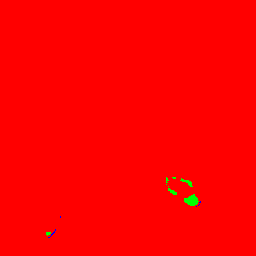}
\includegraphics[width=0.24\textwidth]{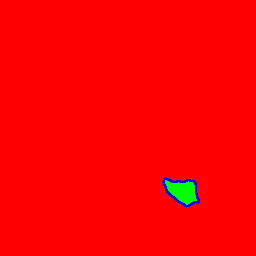}

\caption{Segmentations obtained from the U-Net$_{1}$ (baseline) model (\emph{first}), DDT + U-Net$_{1}$ (\emph{second}), DDT + U-Net$_{2}$ (\emph{third}) and \ac{BTGT} (\emph{fourth}). The DDT + U-Net$_{2}$ outperforms the others due to the reduced number of false positives compared with the U-Net$_{1}$ (baseline) and DDT + U-Net$_{1}$ models, respectively.}
\label{fig:single-cell-pred}
\end{figure}

The proposed method outperforms the UNet$_{1}$ (baseline), particularly when border prediction metrics (\emph{e.g.,} \ac{BDE} and the \ac{WDMC} ) are computed. Both metrics are more relevant and appropriate to measure segmentation accuracy, as tracking methods usually calculate the centroid based on the object edges, and also post-processing effects can correct any \textit{holes} in the object of interest. Using the Dice coefficient or the F1-score gives th same weight to all the pixels, which can give misleading interpretations of the image segmenatation performance. However, the \ac{DDT} output seems to be very noise sensitive (see Figure \ref{fig:Back-Model-Noise}). The \ac{DDT} often amplifies noise when few cells are present, thus hindering the top-model predictions (see Figure \ref{fig:single-cell-pred}). These faulty predictions due to the amplified noise might degrade the benefit of including morphological information.

% Figure 6

Adding the original image to the top model along with the inverse distance transform seems to improve the \ac{DDT} output accuracy. Moreover, it also boosts the overall model performance thanks to the texture information provided. Our results also suggest the possibility to include the \ac{DDT} output as a post-processing in order to remove artifacts. The Wilcoxon statistical test was critical to properly analyse the significance of the obtained results.

% Figure 7

The injection of morphological knowledge can benefit deep \ac{CNN} architectures to enhance instance segmentation accuracy \cite{decenciere2018dealing}. Differently from \cite{guerrero2018multiclass}, in which morphological information was used to weight the loss function, our model still remains simple and easy to train. The proposed back-bone model learns the inverse distance transform by using pixel-wise labeling and adding instance-wise morphological information. The work in \cite{decenciere2018dealing} utilized simple and traditional pre-processing methods.

The behavior of other \ac{CNN}s \cite{he2017mask} after injecting morphological information will be addressed in the future. Indeed, additional experiments with different datasets and \ac{CNN}s are required to provide further evidence regarding the advantages of enriching the image with instance-wise morphological information. To reduce the labeling dependency, semi-supervised learning \cite{papandreou2015weakly}, data augmentation techniques and generative models \cite{sixt2018rendergan} will be employed to enlarge our dataset and explore the advantages of adding morphological information.

% \bibliographystyle{spbasic}
% \addcontentsline{toc}{section}{\refname}\bibliography{biblio}

\bibliographystyle{splncs04}

\end{document}